\documentclass[10pt,twocolumn]{article}
\usepackage[fontsize=9.25pt]{fontsize}

\usepackage[letterpaper,margin=0.75in,columnsep=0.24in]{geometry}
\usepackage{amsmath}                 
\usepackage{newtxtext,newtxmath}     
\usepackage{booktabs}
\usepackage{graphicx}
\usepackage{multirow}
\usepackage{subcaption}
\usepackage[hidelinks]{hyperref}
\usepackage[labelfont=bf,font=small,skip=4pt]{caption}
\usepackage{microtype}
\usepackage{balance}

\newcommand\blfootnote[1]{%
  \begingroup
  \renewcommand\thefootnote{}\footnote{#1}%
  \addtocounter{footnote}{-1}%
  \endgroup
}

\usepackage{titlesec}
\titlespacing*{\section}{0pt}{0.9ex plus .2ex}{0.5ex}
\titlespacing*{\subsection}{0pt}{0.7ex plus .2ex}{0.4ex}
\titleformat{\section}{\normalfont\large\bfseries}{\thesection}{0.6em}{}
\titleformat{\subsection}{\normalfont\normalsize\bfseries}{\thesubsection}{0.5em}{}

\title{\bf Ternary Vision Encoders Without the Straight-Through\\ Estimator: Toward Onboard Robot Perception}
\author{%
  Akshat Bhalani\\[2pt]
  \normalsize Dwarkadas J. Sanghvi College of Engineering, Mumbai, India\\
  \normalsize\texttt{akshat.bhalani85@svkmmumbai.onmicrosoft.com}%
}
\date{}

\begin{document}
\twocolumn[
  \begin{@twocolumnfalse}
  \maketitle
  \vspace{-2.4em}
  \begin{abstract}
  \noindent\small
  Ternary policies such as BitVLA quantize the language backbone to 1.58 bits,
  after which the vision encoder, roughly 0.6\,B parameters in OpenVLA, becomes a
  leading term in what onboard memory remains. Ternary weights $\{-1,0,+1\}$ pack
  into 2 bits, so a ternary linear layer is at most $16\times$ smaller than its
  FP32 counterpart. BitVLA reaches this operating point using a straight-through
  estimator (STE) for the gradient and a pretrained full-precision teacher for the
  vision tower. We study whether either is necessary. Training three backbones
  from random initialization under one matched recipe, an STE-free sigmoid
  relaxation we call NoSTE matches STE-based training on both transformer
  encoders, within our seed standard deviation on ViT-7/CIFAR-10 and Swin-T/Tiny-ImageNet. Training does retain a short-lived full-precision
  target, a frozen copy of each stage discarded shortly after that stage is
  quantized, but no pretrained teacher. We then identify why the standard
  correction for this relaxation's known weakness, sigmoid saturation, fails: the floor required to keep it finite fixes the effective gradient at
  roughly $100\times$ that of STE across the saturated region, reducing the
  estimator to a badly scaled STE. An ablation attributes the result to the
  quantization schedule rather than to the gradient estimator. Finally, Swin-T's
  three patch-merging projections, 5.6\% of parameters and left in full precision
  by standard recipes, are shown to occupy nearly half the packed model, bounding
  achievable compression at $8.2\times$ where ternarizing them would permit
  $14.2\times$.Code and training configurations will be released
publicly soon.
  \end{abstract}
  \vspace{0.7em}
  \end{@twocolumnfalse}
]

\section{Introduction}
\blfootnote{Accepted at the Workshop on Efficient Foundation
Models for Real-Time Embodied AI (EEAI), Conference on Robot Learning (CoRL). Non-archival.}

A vision-language-action policy must satisfy a fixed power and memory budget on
the robot rather than merely run on a datacenter GPU. Its visual front end is a
ViT-family encoder, roughly 0.6\,B parameters against a 7\,B language backbone in
OpenVLA~\cite{openvla}. That is a minority of the policy, but a leading term in
what remains once the backbone is itself ternarized, and so a compression target
in its own right. Ternary weights carry 1.58 bits and pack into 2, so a ternary
linear layer is at most $16\times$ smaller than FP32, and every compression figure
below is quoted against that 2-bit accounting. They also replace multiplication
with addition and sign flips, and leave exact zeros, the sparsity, that a sparse
kernel can skip.

BitVLA~\cite{bitvla} has shown this operating point to be practical, reporting a
1.58-bit policy that runs in approximately 1.4\,GB, $11.0\times$ less memory and
$4.4\times$ lower end-to-end latency than OpenVLA-OFT~\cite{openvla,oft}. It
relies on two mechanisms: a language backbone built on
BitNet~b1.58~\cite{bitnet}, whose quantization-aware training propagates gradients
through the ternary rounding with the STE~\cite{bengio2013}, and a vision tower
produced by quantize-then-distill, which compresses a full-precision encoder under
a full-precision teacher.

Both carry a cost. The STE assumes $\partial Q/\partial w = 1$, so what it
propagates is not the gradient of any smooth objective, and the resulting bias is
known to destabilize training, most severely in attention
layers~\cite{frumkin2023,liu2023osc}. A pretrained teacher presumes a checkpoint
in the target regime, which robot-specific perception modules frequently lack.
This paper asks whether a ternary vision encoder can be trained without either,
from random initialization, at no cost in accuracy.

We study it on image classification rather than on manipulation policies: the
object of study is the training dynamics of from-scratch ternary optimization,
which isolate more cleanly at this scale than inside a policy. This is a
prerequisite to ternarizing a VLA encoder, not a substitute for it.

\textbf{Contributions.} We call this recipe NoSTE. We
show that it matches STE-based training on two transformer encoders trained from
scratch (Section~\ref{sec:main}); that the standard saturation correction cannot
succeed with a constant gradient floor, which sets the amplification rather than
bounding it (Section~\ref{sec:resc}); and that the patch-merging projections,
routinely left in full precision, dominate the compressed footprint of a
hierarchical encoder (Section~\ref{sec:mem}).

\section{Related Work}

TWN~\cite{twn} established the dominant ternary recipe, a threshold at
$0.7\cdot\text{mean}|w|$ with a closed-form scale and an STE backward pass, and
TTQ~\cite{ttq} extended it with learned asymmetric levels.
BitNet~b1.58~\cite{bitnet} scaled ternary weights to language models trained from
scratch, and BitVLA~\cite{bitvla} carried the operating point to manipulation
policies; all four rely on the STE~\cite{bengio2013}, as does LSQ~\cite{lsq},
which learns the quantizer step size directly. Schoenbauer et
al.~\cite{schoenbauer2024} show that a broad class of custom weight-gradient
estimators is equivalent to the STE up to a change in learning rate and
initialization, a result that bears directly on Section~\ref{sec:resc}.
Relaxation-based alternatives to the STE predate this work: relaxed
quantization~\cite{louizos2019} anneals a categorical distribution over the
quantization grid, and ProxQuant~\cite{proxquant} substitutes proximal steps on a
regularized objective; both are evaluated on CNNs and LSTMs rather than on
vision transformers trained from scratch. Ye et al.~\cite{ye2026} instead
replace the STE with a denoising dequantization derived
from a ridge-regression objective, reporting strong results on pretrained language
models; we reproduce it as a baseline. Low-bit vision transformers are separately
known to be sensitive to discretization in the attention
projections~\cite{frumkin2023,qvit}, with weight oscillation a principal failure
mode~\cite{liu2023osc}, which is why our threshold schedule treats attention and
MLP layers asymmetrically. Progressive precision reduction is established for
CNNs~\cite{zhuang2021}, and TerViT~\cite{tervit} brings it to vision transformers
by training an 8-bit model and using it to initialize the ternary one; we stage by
depth rather than bit-width, and from random initialization rather than a
quantized proxy. FTerViT~\cite{ftervit} removes the full-precision remainder that
Section~\ref{sec:mem} quantifies, ternarizing the patch embedding, LayerNorm and
head of columnar DeiT backbones under distillation; binarization pushes the same
backbones to one bit~\cite{bivit}. Patch merging has no
counterpart in a columnar backbone, and our contribution there is the accounting
rather than the removal.

\section{Method}

\subsection{Soft ternary projection}

We maintain a real-valued latent weight $w$ and project it softly to ternary in
the forward pass. With per-layer threshold $\delta$, temperature $t$, and a
learnable per-output-row scale $\alpha$,
\begin{equation}
\begin{aligned}
\tilde w_{\text{soft}} &= \Big[\sigma\big(\tfrac{w-\delta}{t}\big)
 - \sigma\big(\tfrac{-w-\delta}{t}\big)\Big]\alpha,\\[1pt]
\tilde w_{\text{hard}} &= \operatorname{sign}(w)\,\mathbf{1}\big[|w|{>}\delta\big]\,\alpha .
\end{aligned}
\label{eq:soft}
\end{equation}
The two sigmoids form a smooth three-level staircase in which weights well inside
$[-\delta,\delta]$ map toward zero and weights well outside map toward
$\pm\alpha$; $t$ sets the width of the transition. As $t \to 0$ the soft weight converges
to the hard projection. Evaluation always uses $\tilde w_{\text{hard}}$, so no
reported result depends on the relaxation being present at inference.

Gradients flow through Eq.~\ref{eq:soft} itself. No STE appears anywhere in the
training graph; the backward pass is the exact derivative of the forward function.

\subsection{Threshold and scale management}

The threshold is a tracked statistic rather than a fixed hyperparameter. Every
five epochs it is updated by an exponential moving average toward a multiple of
the layer's mean magnitude, $\delta \leftarrow 0.7\,\delta +
0.3\,c\cdot\text{mean}|w|$, floored at $0.04$ so that a collapsing layer cannot
drive its own dead zone to zero. The coefficient $c$ is $0.55$ for QKV projections
and $0.85$ for MLP layers: attention is the more sensitive consumer of
discretization, so QKV receives a narrower dead zone and retains more non-zero
weights. This asymmetry is the principal architectural prior in the method and is
directly visible in the final sparsity profile (Section~\ref{sec:mem}).

The scale $\alpha$ is learned per output row, initialized to
$\|w_{i\cdot}\|_2/\sqrt{\text{fan\_in}}$, and given its own optimizer group with
zero weight decay so that it can grow to absorb the magnitude the projection
discards. It is the only parameter able to compensate for the quantizer.

\subsection{Auxiliary objectives}

Three auxiliary losses act on the soft weights, each targeting a distinct failure
mode of the relaxation.

\textbf{Gated commitment.} We penalize $\|\tilde w_{\text{soft}} - \tilde
w_{\text{hard}}\|^2$ to draw soft weights onto the discrete levels, weighted per
layer by $\mathbf{1}\big[g{>}\tau_{\text{gate}}\big]\,(g/\tau_{\text{gate}})$ with $g = \text{mean}|\tilde
w_{\text{soft}} - \tilde w_{\text{hard}}|$ and $\tau_{\text{gate}} = 0.08$. Layers
whose gap has fallen below $\tau_{\text{gate}}$ leave the loss entirely, which
avoids over-regularizing
converged layers while the remainder of the network is still moving.

\textbf{Discrete consistency.} Every $K{=}4$ steps we run a hard-weight forward
pass under \texttt{no\_grad} and add a KL divergence ($T{=}4$) between the soft-
and hard-weight logits, with gradient flowing only through the soft branch. This
addresses the failure mode in which the soft model trains well but the deployed
discrete model does not track it, which a training curve does not reveal.

\textbf{Magnitude penalty.} A hinge on $|w| > 2\delta$ keeps latent weights out of
deep saturation, where the derivative of Eq.~\ref{eq:soft} is numerically
negligible (Section~\ref{sec:resc}).

The total objective is $\mathcal{L} = \mathcal{L}_{\text{task}} +
\lambda_c\mathcal{L}_{\text{commit}} + \lambda_h\mathcal{L}_{\text{hard}} +
\lambda_m\mathcal{L}_{\text{mag}}$, with all coefficients ramped linearly from
zero over the annealing phase (Section~\ref{sec:sched}) so that early training
follows the task loss alone.

\subsection{Progressive quantization schedule}
\label{sec:sched}

Quantizing every layer at once from random initialization deprives the network of
clean gradient during the epochs in which it forms its representations. Staging is
also what makes the rest of the schedule usable: staging and the final temperature
anneal together are worth 9.9\,pp (Table~\ref{tab:ablation}). We instead quantize
by stage on a fixed schedule. Swin-T trains in FP32 to epoch 80, then quantizes
\texttt{layers.3} at 80 and \texttt{layers.2--3} at 140 (both $t{=}0.5$) and all
stages at 200, annealing $t$ to 0.004; from epoch 370 it hardens at $t{=}0.002$
under the commitment loss alone.

Three implementation details govern whether the transitions are survivable. The
temperature is global rather than per layer: a per-layer schedule leaves
previously quantized layers cold while a new stage is warm, and those layers
cannot then adapt to the new stage's activation statistics. At each transition
we retain a frozen FP32 copy of the newly quantized layers for 25 epochs, capture each ternary layer's
input with a forward hook, run the frozen copy on that input, and add an MSE loss
between the two outputs with a weight decaying linearly to zero. This is
self-distillation from the model's own weights at the transition: the copy is made
at the moment of quantization and discarded 25 epochs later, so no pretrained
checkpoint is involved. It is nonetheless a full-precision target, and we do not
ablate its contribution. Finally, new parameter groups must be added to the live optimizer
via \texttt{add\_param\_group()} rather than rebuilding it, since rebuilding
discards cosine-schedule position and Adam moments and costs 0.4--0.8\,pp.

\begin{figure*}[t]
\centering
\begin{subfigure}[t]{0.485\textwidth}
  \includegraphics[width=0.84\linewidth]{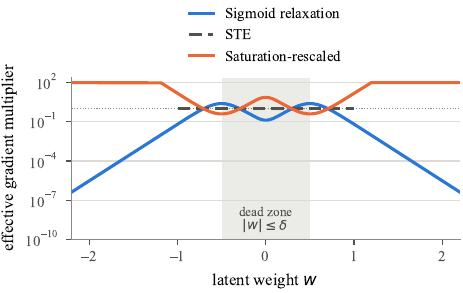}
  \caption{Gradient multiplier at $\delta{=}0.5$, $t{=}0.1$. Sigmoid relaxation
  attenuates to $\approx\!10^{-9}$ at $w{=}2$; a floor $\epsilon{=}10^{-2}$
  overshoots the STE multiplier by $100\times$ across the saturated region.}
  \label{fig:sat}
\end{subfigure}\hfill
\begin{subfigure}[t]{0.485\textwidth}
  \includegraphics[width=0.84\linewidth]{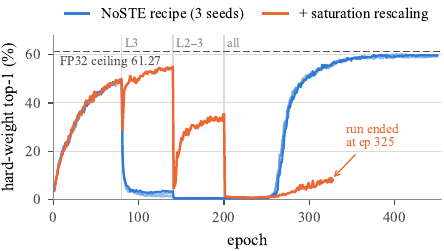}
  \caption{Swin-T / Tiny-ImageNet, hard-weight top-1 per epoch; grey verticals
  mark stage transitions. NoSTE recovers from each; with saturation rescaling it
  degrades at the second and never recovers once fully ternary.}
  \label{fig:curves}
\end{subfigure}
\caption{Saturation rescaling: why inverting the attenuation cannot work (a),
and what that failure looks like in training (b).}
\end{figure*}

\section{Experimental Setup}

We train ResNet-18 and a compact ViT-7 (depth 7, width 384,
$\text{mlp\_ratio}{=}3$) on CIFAR-10, and Swin-T on Tiny-ImageNet at
$64{\times}64$ with window 4, all from random initialization. Patch embedding,
normalization and the classifier remain in FP32; all attention and MLP linear
layers are ternary. Optimization uses AdamW with three parameter groups, covering
FP32 weights ($\text{wd}{=}0.05$), latent weights ($\lambda_{\text{wd}}{=}5{\times}10^{-4}$)
and scales ($\text{wd}{=}0$), at learning rate $5{\times}10^{-4}$ with cosine
annealing and a 20-epoch warmup. Swin-T runs 450 epochs at batch 256 with
RandomResizedCrop, ColorJitter, Mixup, CutMix and RandomErasing. The Swin-T and
ViT-7 runs use a single NVIDIA H100 80\,GB, taking approximately 40\,h (450 epochs) and
6\,h (320 epochs); the ResNet-18 runs use a single T4 and take approximately 4\,h
(250 epochs). Those times include the auxiliary work: a no-grad forward every
four steps for the discrete-consistency loss, and, inside each 25-epoch
post-transition window, a forward through the frozen copy of the new stage.

Baselines are TWN~\cite{twn}, which uses the STE, and Ye et al.~\cite{ye2026},
which does not. Both are re-run under this recipe with identical augmentation,
optimizer, epoch budget and $\lambda_{\text{wd}}$. Our results use seeds $\{42,123,7\}$
throughout; baseline seed coverage varies with the compute each run required and
is given per entry in Table~\ref{tab:main}.

\section{Results}
\label{sec:main}

Table~\ref{tab:main} reports the main comparison. On both transformer backbones,
the architectures relevant to VLA encoders, removing the STE costs no measurable
accuracy. On ViT-7 the two methods fall within 0.35\,pp of each other at
comparable sparsity, inside the $\pm0.39$ seed variation of our own runs. On
Swin-T, the most difficult task considered here and the only one with 200 classes,
NoSTE reaches 59.90\% at 47.7\% sparsity against 59.36\% at 49.5\%. That 0.54\,pp
is smaller than our own seed standard deviation, so we read it as a match rather
than a win; it closes to within 1.37\,pp of the FP32 ceiling.

ResNet-18 is the exception, where the STE remains ahead by approximately 1\,pp. We
have no controlled explanation for that gap and do not claim parity on
convolutional backbones.

The STE-free baseline of Ye et al.\ is strongly regime-dependent. On ViT-7 it
reaches only 67.38\%, driven by its per-row scale $s_f = \max_j |w_{ij}|$. For a
row of $n$ i.i.d.\ Gaussian weights, extreme-value theory gives $\max|w| \approx
\sqrt{2\ln n}\,\sigma$, which is $3.45\sigma$ asymptotically and nearer $3\sigma$
at ViT-7's finite width $n{=}384$. The dead zone therefore covers most of the row
near initialization, and the QKV projections never form. Progressive
staging largely recovers the method on Swin-T, at 59.40\%, because by then each
stage's weights carry 80 to 200 epochs of gradient history and are no longer
Gaussian. The original work, validated on pretrained language models, had no
occasion to surface this.

\begin{table}[!tb]
\centering\small
\setlength{\tabcolsep}{4pt}
\begin{tabular}{llcrrc}
\toprule
Backbone / data & Method & Grad. & Top-1 & Spars. & $n$\\
\midrule
\multirow{4}{*}{\shortstack[l]{ResNet-18\\CIFAR-10}}
 & FP32      & ---    & 96.64 & 0.0 & 1\\
 & TWN       & STE    & \textbf{96.10}\,$\pm$\,0.03 & 51.5 & 2\\
 & Ye et al. & no STE & 93.98\,$\pm$\,0.18 & 91.5 & 3\\
 & NoSTE     & no STE & 95.02\,$\pm$\,0.02 & 47.5 & 3\\
\midrule
\multirow{4}{*}{\shortstack[l]{ViT-7\\CIFAR-10}}
 & FP32      & ---    & 92.10 & 0.0 & 1\\
 & TWN       & STE    & \textbf{90.80} & 43.3 & 1\\
 & Ye et al. & no STE & 67.38 & 79.6 & 1\\
 & NoSTE     & no STE & 90.45\,$\pm$\,0.39 & 45.3 & 3\\
\midrule
\multirow{4}{*}{\shortstack[l]{Swin-T\\Tiny-ImageNet}}
 & FP32      & ---    & 61.27 & 0.0 & 1\\
 & TWN       & STE    & 59.36 & 49.5 & 1\\
 & Ye et al. & no STE & 59.40 & 64.2 & 1\\
 & NoSTE     & no STE & \textbf{59.90}\,$\pm$\,0.59 & 47.7 & 3\\
\bottomrule
\end{tabular}
\caption{From-scratch ternary training: hard-weight top-1 (\%), sparsity (the
fraction of ternary weights at exactly zero, \%), and the number of seeds $n$ over
which each entry is a mean\,$\pm$\,sample standard deviation. Sparsity is a
property of the solution rather than a target.}
\label{tab:main}
\end{table}

\subsection{Ablation study}

Table~\ref{tab:ablation} attributes the final accuracy across the components of
NoSTE. The ordering is informative: the components that matter control when
and how quickly weights discretize rather than reshaping the gradient.

On Swin-T the schedule accounts for almost all of the gain. Staging alone adds
$1.44$\,pp over quantizing every layer at once, and the final temperature anneal
and hardening phase, which becomes usable only once the stages are separated, adds
a further $7.87$\,pp; latent weight decay then adds $0.61$\,pp on top. On
ResNet-18 each component was instead run as a single change against one shared
baseline, so those deltas do not compose. There latent weight decay,
$\lambda_{\text{wd}}{=}5{\times}10^{-4}$, is the largest single contributor at
$+1.01$\,pp, ahead of the faster EMA threshold at $+0.92$\,pp. It holds latent
weights near the threshold rather than letting them drift into deep saturation,
which stabilizes the threshold estimate and shapes the sparsity structure of
Section~\ref{sec:mem}. Removing the magnitude penalty on that backbone moves
accuracy by $+0.14$\,pp, so that term does no measurable work there.
Optimizer-state continuity accounts for a further 0.4--0.8\,pp.

For the discrete-consistency loss we report the quantity it controls: at
convergence the hard models fall within 0.6 to 1.2\,pp of the soft models they
were trained as, a mean of 0.95\,pp over three seeds. The controlled removal has
not been run, so this is an achieved gap rather than an attributed delta. Neither
is the stage-wise self-distillation of Section~\ref{sec:sched} removed in
isolation, and three of the Swin-T rows are single runs. Table~\ref{tab:ablation}
therefore establishes a working combination more firmly than it isolates every
component of it.

\begin{table}[!tb]
\centering\small
\setlength{\tabcolsep}{4pt}
\begin{tabular}{lcc}
\toprule
Configuration & Top-1 & Spars.\\
\midrule
\multicolumn{3}{l}{\emph{Swin-T / Tiny-ImageNet} (cumulative)}\\
All layers ternary at once        & 49.98 & 51.6\\
\;+ progressive staging           & 51.42 & 50.8\\
\;+ final temperature anneal, harden & 59.29 & 48.1\\
\;+ latent wd $5{\times}10^{-4}$  & \textbf{59.90} & 47.7\\
\;+ saturation rescaling          & \multicolumn{2}{c}{\emph{no convergence}}\\
\midrule
\multicolumn{3}{l}{\emph{ResNet-18 / CIFAR-10} (one change each)}\\
Clean baseline                    & 94.17 & 41.9\\
\;faster EMA threshold            & 95.09 & 48.0\\
\;latent wd $5{\times}10^{-4}$    & \textbf{95.14} & 47.9\\
\;magnitude penalty removed       & 94.27 & 41.8\\
\bottomrule
\end{tabular}
\caption{Component study. Swin-T rows are cumulative; ResNet-18 rows are
independent single-parameter changes against one baseline, so their deltas do not
compose. Swin-T rows are single runs except 59.29 and 59.90, three-seed means
($\pm0.08$, $\pm0.59$).}
\label{tab:ablation}
\end{table}

\section{Analysis}

\subsection{Saturation rescaling and its gradient floor}
\label{sec:resc}

Differentiating Eq.~\ref{eq:soft} gives $\partial \tilde w_{\text{soft}}/\partial
w = [\sigma'(\tfrac{w-\delta}{t}) + \sigma'(\tfrac{-w-\delta}{t})]/t$, which
vanishes once a weight sits far outside the dead zone. At $w{=}2$, $\delta{=}0.5$
and $t{=}0.1$ it evaluates to approximately $10^{-9}$, so a confidently assigned
weight receives effectively no gradient regardless of the task loss. The practical
advantage of the STE is precisely that it sidesteps this by ignoring the
quantizer.

The established repair inverts the attenuation using the same derivative, dividing
rather than multiplying in the backward pass, with a floor $\epsilon$ for
numerical stability:
\begin{equation}
\hat g_w = g_{\tilde w} \big/ \max\!\big(\big|\partial \tilde w_{\text{soft}}/\partial w\big|,\ \epsilon\big).
\label{eq:rescale}
\end{equation}
The correction derives from the same relaxation used in the forward pass, and it
does restore full-magnitude gradients to frozen weights.

Fig.~\ref{fig:sat} shows why that is insufficient. Across the saturated region the
floor does not bound the amplification but sets it, because the true derivative
there lies orders of magnitude below $\epsilon$. Every confidently assigned weight
therefore receives a gradient $1/\epsilon = 100\times$ larger than the STE would
supply, applied uniformly to exactly those weights that are most committed.

The measurements follow the analysis. Across ten runs spanning $\epsilon \in
\{10^{-3}, 10^{-2}\}$ and two architectures, the correction never produced a
competitive model. At $\epsilon{=}10^{-3}$ training diverges within 80 epochs. At
$\epsilon{=}10^{-2}$ the behaviour matches a uniform overshoot: tolerable while
one stage is quantized, destructive once all are. The Swin-T trajectory in
Fig.~\ref{fig:curves} peaks at 55.07\% and sits at 8\% after 125 all-ternary
epochs, by which point the same recipe without rescaling has returned to 60\%; the
one ViT-7 run that completed scored 80.62\% against 90.45\%.

Inverting the attenuation is therefore the wrong remedy, and this is a concrete
instance of the equivalence reported by Schoenbauer
et al.~\cite{schoenbauer2024}. Over the saturated region Eq.~\ref{eq:rescale} is
not an alternative to the STE but the STE rescaled by $1/\epsilon$, inheriting its
instability without its calibration. This bounds one correction and not a class: a per-layer
floor adapted to the observed derivative distribution would not share the
property, and nothing here rules out gradient corrections in general.

\begin{figure}[!tb]
\centering
\includegraphics[width=0.82\linewidth]{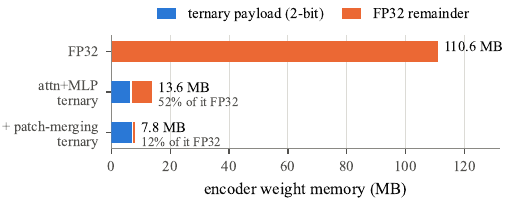}
\caption{Swin-T encoder weight memory. Ternary weights packed at 2 bits, all other
parameters FP32; analytical parameter counts.}
\label{fig:mem}
\end{figure}

\subsection{Encoder memory accounting}
\label{sec:mem}

Counting attention and MLP linear weights at 2 bits and all other parameters at
FP32, ViT-7 falls from 41.6\,MB to 2.90\,MB, a factor of $14.3\times$ and close to
the $16\times$ usually quoted. Swin-T falls from 110.6\,MB to 13.57\,MB, only
$8.2\times$.

The difference lies in the full-precision remainder, which is concentrated rather
than diffuse. Swin-T retains 1.78\,M parameters in full precision, 6.4\% of the
model, and they occupy 7.10\,MB, or 52\% of the packed encoder
(Fig.~\ref{fig:mem}). Almost all of this is a single component: the three
patch-merging projections that halve spatial resolution between stages hold
1.55\,M parameters and 6.19\,MB, which is 5.6\% of the parameters and 46\% of the
compressed model. They are ordinary linear layers, as amenable to ternarization as
any other, but fall outside the \texttt{attn.*} and \texttt{mlp.*} name patterns
that ternary recipes quantize, including ours.

Were they also ternarized, the encoder would fall to 7.77\,MB and compression
would rise to $14.2\times$, with the full-precision share falling to 12\%. We have
not trained a model in that configuration, so this is the compression the
remaining full-precision parameters forgo, not a measured result; that comparable
components tolerate ternarization elsewhere~\cite{ftervit} makes it plausible
rather than established. The accounting generalizes: once attention and MLP are
ternary, the achieved ratio is set by whatever architecture-specific components
remain, and in hierarchical backbones those are not small.

Sparsity is also structured by depth rather than uniform, which bears on whether a
sparse kernel can exploit it. First-stage MLP layers reach 72 to 75\% sparsity
while middle and late stages settle at 49 to 56\%, a profile that appears only
with weight decay on the latent weights. QKV projections remain near 30\%
throughout, a direct imprint of the $c{=}0.55$ coefficient. By contrast the
uniformly high sparsity Ye et al.\ produce on ViT-7, at 79.6\%, comes at
substantial accuracy cost, which is why sparsity is reported here only alongside
its accuracy.

\section{Limitations}

Several limitations bound these results. The evaluation is image classification
rather than closed-loop control, so we report top-1 accuracy and not task success.
Every model is trained from scratch, so nothing here speaks to whether
ternarization preserves pretrained visual features. The memory figures are
analytical parameter counts: we serialize no packed model, implement no ternary
kernel, and measure no inference latency or deployed peak memory. What the
evidence supports is training methodology and storage accounting, not onboard
acceleration or realized sparse-kernel savings.

Scale is the next. The largest encoder studied here is Swin-T, 110.6\,MB in FP32
and more than an order of magnitude smaller than the 0.6\,B-parameter towers the
opening argument concerns, trained at $64{\times}64$ rather than at ImageNet
resolution. Nothing in the schedule is specific to small models, but whether a
stage-wise anneal stays stable at that width and depth is untested, and the stage
boundaries are set by hand rather than derived from any measured property of the
run.

The ablation has a boundary of its own. Table~\ref{tab:ablation} separates the
quantization schedule from the gradient estimator; it does not separate every
auxiliary term. Stage-wise self-distillation and the discrete-consistency loss are
not individually ablated, so their contribution is asserted by construction rather
than measured. The ResNet-18 block
is three single-change runs against one shared baseline rather than a cumulative
ladder, and several Table~\ref{tab:main} entries are single runs, which should be
read against the 0.4 to 0.6\,pp seed spread of the three-seed transformer rows.

\section{Conclusion}

\looseness=-1
Ternary vision encoders can be trained from random initialization without the
straight-through estimator and without a pretrained full-precision teacher,
matching STE-based training on two transformer backbones. Training does use a
full-precision target, a frozen copy of each stage taken as it is quantized and
discarded 25 epochs later; what it does not need is a pretrained checkpoint in the
target regime, the assumption robot-specific perception modules fail. The
ablation is not an argument against going STE-free: it says the STE is unnecessary once the quantization
schedule is right, and that schedule is exactly what makes an exact-gradient
estimator viable. The sharpest practical finding lies elsewhere. Swin-T's
patch-merging projections are 5.6\% of its parameters but 46\% of the packed
model, so leaving them in full precision caps compression at $8.2\times$ against
the $14.2\times$ otherwise available. A ternary encoder budget belongs over the
whole encoder, not its transformer blocks.

Three steps follow directly. A packed ternary kernel and a serialized model would
turn the accounting of Section~\ref{sec:mem} into measured latency and deployed
peak memory, which is what an onboard budget is actually denominated in. Ablating
the self-distillation and discrete-consistency terms would establish how much of
the result the schedule carries on its own. And applying NoSTE to an open VLA's
vision tower~\cite{openvla,smolvla}, then reporting LIBERO success~\cite{libero}
against control frequency, would test the premise the paper opens with.

{\mdseries

}


\begin{thebibliography}{99}
\bibitem{openvla} M.~J.~Kim, K.~Pertsch, S.~Karamcheti, T.~Xiao, A.~Balakrishna,
  S.~Nair, R.~Rafailov, E.~Foster, G.~Lam, P.~Sanketi, Q.~Vuong, T.~Kollar,
  B.~Burchfiel, R.~Tedrake, D.~Sadigh, S.~Levine, P.~Liang, and C.~Finn. OpenVLA:
  An open-source vision-language-action model. \emph{arXiv:2406.09246}, 2024.
\bibitem{bitvla} H.~Wang, C.~Xiong, R.~Wang, and X.~Chen. BitVLA: 1-bit
  vision-language-action models for robotics manipulation.
  \emph{arXiv:2506.07530}, 2025.
\bibitem{oft} M.~J.~Kim, C.~Finn, and P.~Liang. Fine-tuning vision-language-action
  models: Optimizing speed and success. In \emph{Robotics: Science and Systems
  (RSS)}, 2025. \emph{arXiv:2502.19645}.
\bibitem{bitnet} S.~Ma, H.~Wang, L.~Ma, L.~Wang, W.~Wang, S.~Huang, L.~Dong,
  R.~Wang, J.~Xue, and F.~Wei. The era of 1-bit LLMs: All large language models
  are in 1.58 bits. \emph{arXiv:2402.17764}, 2024.
\bibitem{bengio2013} Y.~Bengio, N.~L\'eonard, and A.~Courville. Estimating or
  propagating gradients through stochastic neurons for conditional computation.
  \emph{arXiv:1308.3432}, 2013.
\bibitem{frumkin2023} N.~Frumkin, D.~Gope, and D.~Marculescu. Jumping through
  local minima: Quantization in the loss landscape of vision transformers. In
  \emph{ICCV}, pages 16978--16988, 2023. \emph{arXiv:2308.10814}.
\bibitem{liu2023osc} S.-Y.~Liu, Z.~Liu, and K.-T.~Cheng. Oscillation-free
  quantization for low-bit vision transformers. In \emph{ICML}, PMLR 202, pages
  21813--21824, 2023. \emph{arXiv:2302.02210}.
\bibitem{twn} F.~Li, B.~Liu, X.~Wang, B.~Zhang, and J.~Yan. Ternary weight
  networks. \emph{arXiv:1605.04711}, 2016.
\bibitem{ttq} C.~Zhu, S.~Han, H.~Mao, and W.~J.~Dally. Trained ternary
  quantization. In \emph{ICLR}, 2017. \emph{arXiv:1612.01064}.
\bibitem{lsq} S.~K.~Esser, J.~L.~McKinstry, D.~Bablani, R.~Appuswamy, and
  D.~S.~Modha. Learned step size quantization. In \emph{ICLR}, 2020.
  \emph{arXiv:1902.08153}.
\bibitem{schoenbauer2024} M.~Schoenbauer, D.~Moro, L.~Lew, and A.~Howard. Custom
  gradient estimators are straight-through estimators in disguise.
  \emph{arXiv:2405.05171}, 2024.
\bibitem{louizos2019} C.~Louizos, M.~Reisser, T.~Blankevoort, E.~Gavves, and
  M.~Welling. Relaxed quantization for discretized neural networks. In
  \emph{ICLR}, 2019. \emph{arXiv:1810.01875}.
\bibitem{proxquant} Y.~Bai, Y.-X.~Wang, and E.~Liberty. ProxQuant: Quantized
  neural networks via proximal operators. In \emph{ICLR}, 2019.
  \emph{arXiv:1810.00861}.
\bibitem{ye2026} C.~Ye, G.~Chu, Y.~Liu, Y.~Zhang, L.~Lew, L.~Zhang, M.~Sandler,
  and A.~Howard. Robust training of neural networks at arbitrary precision and
  sparsity. In \emph{ICLR}, 2026. \emph{arXiv:2409.09245}.
\bibitem{qvit} Y.~Li, S.~Xu, B.~Zhang, X.~Cao, P.~Gao, and G.~Guo. Q-ViT:
  Accurate and fully quantized low-bit vision transformer. In \emph{NeurIPS},
  2022. \emph{arXiv:2210.06707}.
\bibitem{zhuang2021} B.~Zhuang, M.~Tan, J.~Liu, L.~Liu, I.~Reid, and C.~Shen.
  Effective training of convolutional neural networks with low-bitwidth weights
  and activations. \emph{IEEE TPAMI}, 44(10):6140--6152, 2021.
\bibitem{tervit} S.~Xu, Y.~Li, T.~Ma, B.~Zeng, B.~Zhang, P.~Gao, and J.~Lv.
  TerViT: An efficient ternary vision transformer. \emph{arXiv:2201.08050}, 2022.
\bibitem{ftervit} S.~Ruci\'nski, P.~Bonazzi, E.~T\"uretken, S.~Narduzzi,
  M.~Magno, and N.~Maamari. FTerViT: Fully ternary vision transformer. In
  \emph{NeurIPS}, 2026. \emph{arXiv:2605.21171}.
\bibitem{bivit} Y.~Li, S.~Xu, M.~Lin, X.~Cao, C.~Liu, X.~Sun, and B.~Zhang.
  Bi-ViT: Pushing the limit of vision transformer quantization.
  \emph{arXiv:2305.12354}, 2023.
\bibitem{smolvla} M.~Shukor, D.~Aubakirova, F.~Capuano, P.~Kooijmans, S.~Palma,
  A.~Zouitine, M.~Aractingi, C.~Pascal, M.~Russi, A.~Marafioti, S.~Alibert,
  M.~Cord, T.~Wolf, and R.~Cadene. SmolVLA: A vision-language-action model for
  affordable and efficient robotics. \emph{arXiv:2506.01844}, 2025.
\bibitem{libero} B.~Liu, Y.~Zhu, C.~Gao, Y.~Feng, Q.~Liu, Y.~Zhu, and P.~Stone.
  LIBERO: Benchmarking knowledge transfer for lifelong robot learning. In
  \emph{NeurIPS Datasets and Benchmarks}, 2023. \emph{arXiv:2306.03310}.
\end{thebibliography}
\end{document}